\documentclass[lettersize,journal]{IEEEtran}
\usepackage{amsmath,amsfonts}
\usepackage{algorithmic}
\usepackage{algorithm}
\usepackage{array}
\usepackage[caption=false,font=normalsize,labelfont=sf,textfont=sf]{subfig}
\usepackage{textcomp}
\usepackage{stfloats}
\usepackage{url}
\usepackage{verbatim}
\usepackage{graphicx}
\usepackage{cite}
\usepackage{url}
\usepackage{subfig}

\begin{document}

\title{RACE-SM: Reinforcement Learning Based Autonomous Control for Social On-Ramp Merging\\}

\author{Jordan Poots
        % <-this % stops a space
\thanks{Jordan Poots is with the School of Mechanical and Aerospace Engineering, Queen's University Belfast, UK (email: jpoots04@qub.ac.uk)}% <-this % stops a space
\thanks{Corresponding author: Jordan Poots}}

\markboth{}%
{Shell \MakeLowercase{\textit{Jordan Poots}}: Reinforcement learning based autonomous control for social on-ramp merging}

\maketitle

\begin{abstract}
Autonomous parallel-style on-ramp merging in human controlled traffic continues to be an existing issue for autonomous vehicle control. Existing non-learning based solutions for vehicle control rely on rules and optimization primarily. These methods have been seen to present significant challenges. Recent advancements in Deep Reinforcement Learning have shown promise and have received significant academic interest however the available learning based approaches show inadequate attention to other highway vehicles and often rely on inaccurate road traffic assumptions. In addition, the parallel-style case is rarely considered. A novel learning based model for acceleration and lane change decision making that explicitly considers the utility to both the ego vehicle and its surrounding vehicles which may be cooperative or uncooperative to produce behaviour that is socially acceptable is proposed. The novel reward function makes use of Social Value Orientation to weight the vehicle's level of social cooperation and is divided into ego vehicle and surrounding vehicle utility which are weighted according to the model’s designated Social Value Orientation. A two-lane highway with an on-ramp divided into a taper-style and parallel-style section is considered. Simulation results indicated the importance of considering surrounding vehicles in reward function design and show that the proposed model matches or surpasses those in literature in terms of collisions while also introducing socially courteous behaviour avoiding near misses and anti-social behaviour through direct consideration of the effect of merging on surrounding vehicles.
\end{abstract}

\begin{IEEEkeywords}
Autonomous driving, on-ramp merging, deep learning, reinforcement learning, social value orientation
\end{IEEEkeywords}

\section{Introduction}
The idea of autonomous vehicles (AVs) has captured both the imagination of sci-fi writers and the minds of engineers for decades. In the modern age, AVs are moving off the screen and onto the streets with several companies trialling solutions. Google's parent company, Alphabet, is offering its Waymo One autonomous taxi service in multiple US locations \cite{Waymo2023}. Robotaxis are also being trialled by Cruise, a subsidiary of General Motors \cite{Cruise2023} and Amazon's Zoox \cite{Zoox2023} among others. 

\begin{figure}[!t]
\centering
\includegraphics[width=\linewidth]{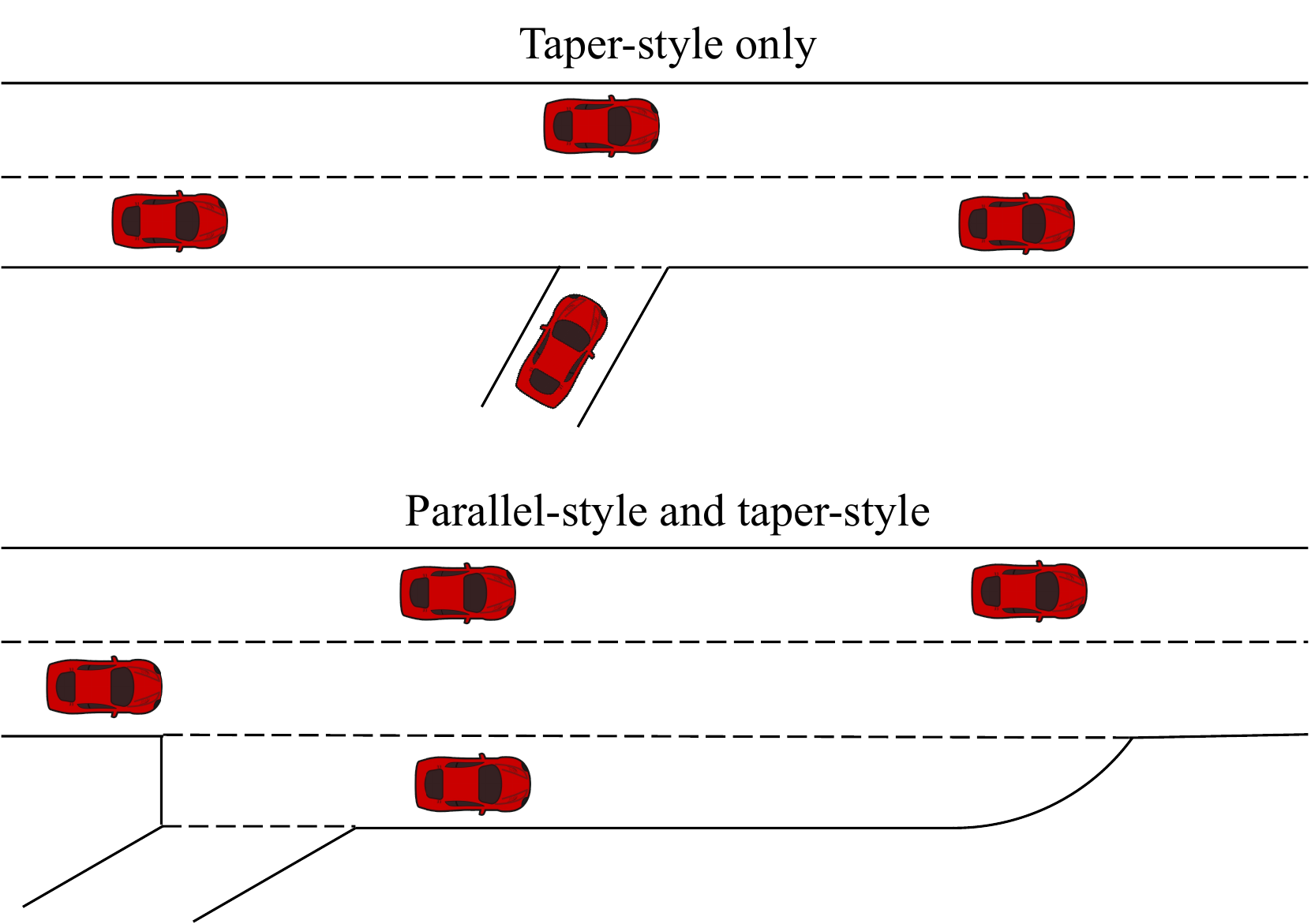}
\caption{A comparison between on-ramps with a taper-style section only and on-ramps with both a taper-style and parallel-style section.}
\label{ramp_types}
\end{figure}

One challenge in vehicle autonomy is decision making for parallel-style on-ramp merging in human controlled traffic. Current predictions indicate that only 50\% of vehicles will be autonomous by 2060 \cite{litman2017autonomous} and thus it is important that any AV be able to navigate human controlled traffic. This challenge involves autonomous decision making on vehicle acceleration and lane changing in response to complex social queues and road conditions. In order to be safe, it is clear any autonomous agent must not just avoid collisions but also be socially courteous, considering its own behaviour and utility as well as that of its surrounding vehicles (SVs) and act in a predictable, human-like manner. Parallel-style ramps feature a merging area within which the ego vehicle could merge at any point, \figurename ~\ref{ramp_types}. This type is more commonly seen in the real world than taper-style only ramps that have a fixed merging point \cite{Wang2021}. The merging area adds an additional layer of complexity to the problem. The sheer number of possibilities and variables along with the social interaction involved makes it difficult to define an optimal one-size-fits-all policy.

Traditionally, rule and optimisation approaches have been proposed to solve this problem, \cite{Zhu2022,torres2017review}. These approaches have suffered from a lack of adaptability and problems in modelling and technical overheads \cite{Wang2021, Liu2022, Ding2020}. Deep Reinforcement Learning (DRL) is a learned approach which makes use of artificial neural networks inspired by the human brain to learn through trial and error in a manner analogous to human learning \cite{sutton2018}, \cite{goodfellow2016}. Due to the learned nature of the models produced by this method, it is well suited to decision making problems which have many possible states. The usage of DRL in the field of autonomous driving is an area of active research and has seen significant academic interest with applications for both detection and control \cite{Muhammad2021}. DRL has already seen extensive usage in on-ramp merging as in \cite{Triest2020, Lin2020, Nie2023} A DRL approach for tackling the problem of parallel-style on-ramp merging will be described. 

Existing DRL methods fail to give sufficient attention in reward function design to the social aspect of merging focusing only on the goals of the ego vehicle and neglecting the effect that a merge might have on surrounding traffic - even if it is successful. Our innovation is the introduction of a novel DRL reward function definition based on the concept of social value orientation (SVO) to produce socially courteous behaviour. This reward function considers explicitly and comprehensively the satisfaction of SVs in human controlled traffic. SVO is a concept from the field of social psychology which describes to what extent a person seeks to maximise or minimise their own and others' outcomes \cite{Griesinger1973} which we use to assign driving behaviour. We demonstrate the importance of SV consideration in reward function design and that direct consideration of SV satisfaction leads to socially courteous behaviour which avoids near misses and anti-social driving behaviour.

First, we define the problem as a Markov Decision Process (MDP) including a novel reward function. Next, we train the agent in a two lane highway populated by medium density traffic which may be cooperative or uncooperative across a range of SVO values. Then, we evaluate test results for the models to demonstrate the requirement of considering SVs in reward function design and the influence of SVO, which is used to produce a range of human-like behaviours within the range of individualist and altruistic. Finally, we evaluate the prosocial model and compare it to other models in the literature. Traditional metrics such as collision rate and average merging velocity are used as well as social metrics such as time to collision (TTC) and gap positioning.

\section{Related Work}
This section is organised as follows. First, heuristic and optimization based approaches will be reviewed. Next, the reinforcement learning based approaches that relate closely to this project will be focused upon. Finally, the relevant literature on the theory and application of social value orientation will be discussed and the reinforcement learning state-of-the-art summarised.

\subsection{Heuristic control}
Traditionally, rule based approaches have been proposed to solve the autonomous driving problem as in \cite{Kachroo1997} and more recently in \cite{Baker2008, Milanes2011, Yang1993}. Baker et al. \cite{Baker2008} employ a slot-based approach for lane merging. Milanes et al. \cite{Milanes2011} apply a Fuzzy logic approach to the on-ramp merging problem. Yang et al. \cite{Yang1993} employ heuristics for longitudinal vehicle control with the goal of gap tracking to a fixed merging point. These approaches choose actions based on pre-defined heuristics. While effective in common scenarios, they have an inability to adapt to complex or unconsidered situations due to their explicit nature \cite{Liu2022}, \cite{Wang2021}. Heuristics are based on experience and intuition and it is difficult to derive heuristics that will apply to all real-world scenarios. Nonetheless, the system described in \cite{Baker2008} is an important piece of AV history as the first winner of the DARPA Urban Challenge. 

\subsection{Optimization approaches}
A number of approaches have been proposed where vehicle control is determined based upon the optimisation of some function. Liu et al. use Model Predictive Control within a game theory framework for AV trajectory planning \cite{Liu2023}. Okuda et al. \cite{Okuda2021} build a driver behaviour model based upon simulated driving data from 28 participants and use Random Model Predictive Control for vehicle control. A further overview of these approaches is provided in \cite{Zhu2022}, \cite{torres2017review}. Optimization approaches often have a low tolerance to unpredictable traffic behaviour due to the assumptions made and high computational overheads limiting their practicality in real world high speed merging scenarios \cite{Wang2021}.

\subsection{Reinforcement leaning approaches}
DRL has a promising ability to learn from experience and successfully navigate less frequent situations. This is an area of active research for vehicle control \cite{Muhammad2021}. DRL has already seen extensive use for the on-ramp merging problem.

Triest et al. \cite{Triest2020} use the A2C algorithm to produce a merging model with a collision rate of 4.2\% within the NGSIM dataset. The control scheme is divided into low and high level and the reward function for the high-level controller considers only vehicle success and longitudinal vehicle progress. The DRL approach proposed by Lin et al. in \cite{Lin2020} for on-ramp merging uses the Deep Deterministic Policy Gradient algorithm to solve the merging problem in a way that minimises the ego vehicle’s jerk. Considered in the reward function was gap positioning, SV speed, trailing vehicle braking and jerk. Nie et al. \cite{Nie2023} make use of PPO algorithm with a reward function focused on the ego vehicle's long-term goals. The authors also introduce a TTC based safety check which replaces inappropriate actions  for increased safety. Wu et al. \cite{Wu2022} use the RTD3 algorithm to train a model using a reward function based on velocity and vehicle state and produce a collision free model when tested in a SUMO simulation populated with a range of driving styles. Wang et al. \cite{Wang2021} use the DQN algorithm for merging in a highway network with a parallel-style merging ramp. Included in the reward function definition are success, safety and efficiency terms. The authors report a collision free model but assume all drivers are uncooperative. 

Lubars et al. \cite{Lubars2021} propose combining model predictive control with DRL to attain a higher level of vehicle safety while attaining the benefits of DRL. The proposed reward function considers success, slow merging and jerk. Testing takes place in a simulated low speed environment using SUMO and attains promising results. Lui et al. \cite{Liu2022} also combine model predictive control with DRL using the SAC-Discrete algorithm and use a more comprehensive reward function. The model was evaluated in a range of traffic conditions.

Within DRL solutions to the on-ramp merging problem, insufficient value is given to the effect on SVs. For example, \cite{Triest2020, Nie2023, Wu2022} considers only the ego vehicles goals while \cite{Lin2020, Liu2022} do not cover SVs comprehensively. In addition, at
times the assumed driving environment is unrealistic when
compared to the real world and the parallel-style on-ramp is
disregarded.

\subsection{Social value orientation}
\begin{figure}[!t]
\centering
\includegraphics[width=\linewidth]{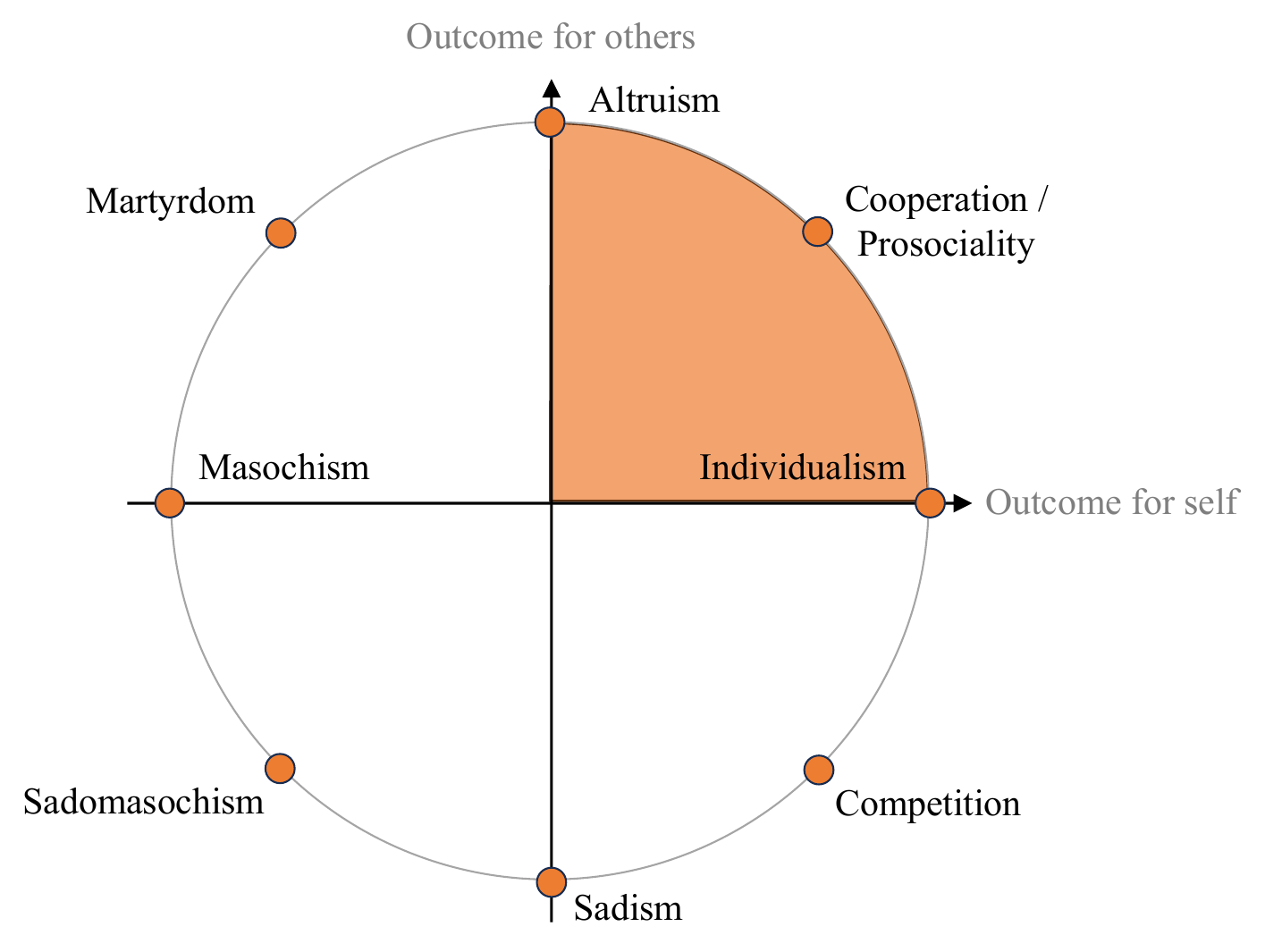}
\caption{The social value orientation ring proposed by Griesinger et al. \cite{Griesinger1973} The highlighted quadrant was used in the reward function design.} \label{svo_ring}
\end{figure}

The concept of social value orientation has helped to categorise human social behaviour. It is an interpersonal trait that indicates an individual's level of preference for self and others in resource allocation and risk-taking behaviours. Griesinger et al. propose a continuous geometric ring model to represent allocation preferences, \figurename \ref{svo_ring} \cite{Griesinger1973}. Within this model, social value orientation is defined as the angle of a straight line from the positive x-axis in a Cartesian coordinate system. In the top right quadrant, self is favour decreasingly as SVO moves towards $\frac{\pi}{2}$.

This concept has already seen use within autonomous driving research. Crosato et al. use SVO in concert with reinforcement learning to vary the attitude of AVs toward pedestrians \cite{crosato2021, crosato2023}. Toghi et al. also use SVO and reinforcement learning to induce sympathetic and cooperative driving behaviour in a multi-agent framework \cite{toghi2021}. Schwarting et al. employ SVO within a game theory approach for AV control and to estimate human vehicle intentions \cite{Schwarting2019}.

\section{Contribution}
A novel reinforcement learning reward function using SVO which considers the goals of the SVs in on-ramp merging explicitly and comprehensively, as well as the goals of the ego vehicle is introduced. This is trained and evaluated for a range of SVOs in a high-speed stochastic environment with both cooperative and uncooperative drivers. Training across a range of SVOs demonstrates the importance of SV consideration in reward function design. It is also demonstrated that within the proposed model, the prosocial SVO value of $\frac{\pi}{4}$ which weights all parties evenly allows for socially courteous vehicle behaviour which responds to cooperative and uncooperative vehicles in a manner which prevents not just collisions but also near misses and road user frustration. This is in contrast to the literature which often fails to appropriately consider the goals of SVs. In addition, this is applied to the lesser studied parallel-style on-ramp case within a single-agent framework.

\section{Background}
\subsection{Reinforcement learning}
In reinforcement learning, an agent learns how to map states to actions to maximise a numerical reward signal. To apply reinforcement learning, the problem is generally modelled as a Markov decision process (MDP). An MDP is defined by the tuple $M = (S, A, P, r, \gamma)$ where $S$ is the set of possible states, $A$ is the set of possible actions, $P$ is a transition probability function, $r$ is a reward function and $\gamma$ is the discount factor. Within an MDP, it is assumed that all information required to make an optimal decision is provided by the environment.

The goal of a reinforcement learning algorithm is to find an optimal policy which maps actions to states such that the expected sum of discounted rewards is maximised for all states. This policy is the optimal solution of the MDP. States and rewards are passed to the agent from the environment and the agent chooses actions which affect the environment. The agent's policy is optimised based on this data.

The expected sum of discounted rewards for a state, $V(s)$, is the total reward the agent expects to receive over the course of an episode when following a policy $\pi$, from the state $s$, at timestep $t$, with future rewards being discounted by $\gamma$ so as to be less valuable to the agent. $V(s)$ is defined by (\ref{value_function}). This is known as the value function. 

\begin{equation}
\label{value_function}
V_{\pi}(s) = \mathbb{E}_{\pi} \left[\sum_{k=0}^{\infty} \gamma^{k}r_{t+k+1} | s_{t} = s\right]
\end{equation}

\begin{figure}[!t]
\centering
\includegraphics[width=\linewidth]{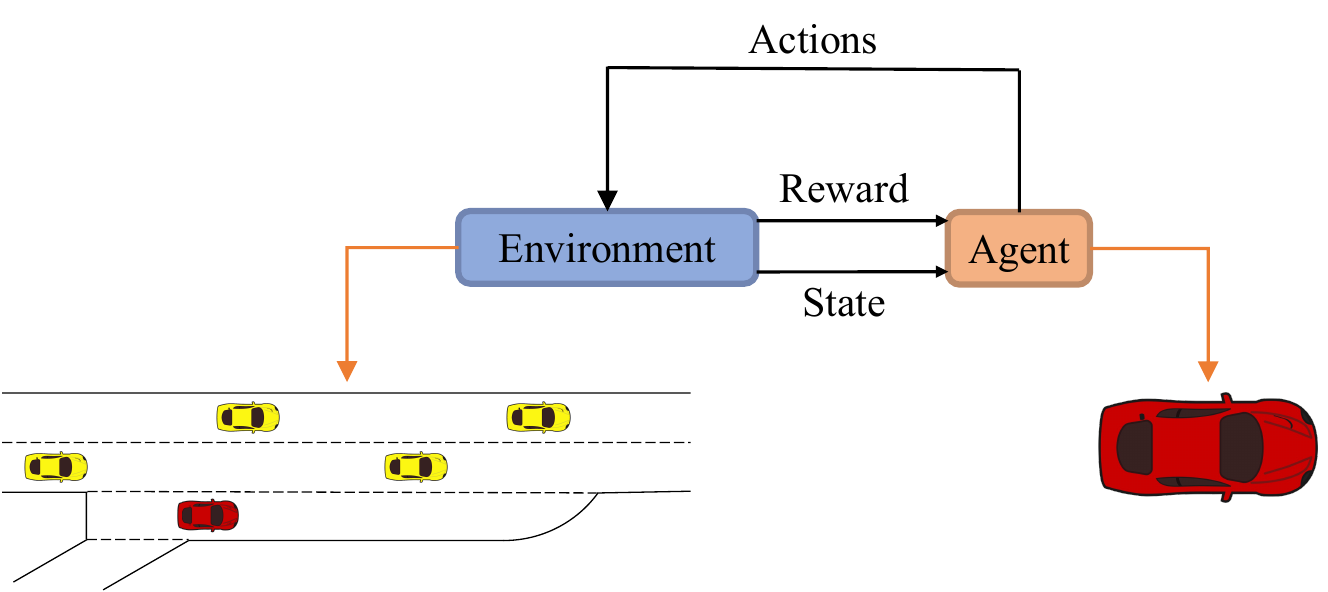}
\caption{Reinforcement learning training cycle visualisation. The environment and agent are both indicated on the diagram.} \label{drl_visualisation}
\end{figure}

The learning process in the context of automated merging is visualised in \figurename ~\ref{drl_visualisation}.

\subsection{Proximal policy optimisation}
The proximal policy optimization algorithms are a group of state-of-the-art reinforcement learning algorithms. Within this report, PPO will refer to the canonical actor critic style PPO-clip algorithm described in \cite{Schulman2017}.

The Q-value of a state action pair is the expected discounted sum of rewards when an action $a$ is taken in state $s$ at timestep $t$ and the policy $\pi$ is followed thereafter. This is defined in (\ref{q_value}).

\begin{equation}
\label{q_value}
Q_{\pi}(s,a) = \mathbb{E}_{\pi} \left[\sum_{k=0}^{\infty} \gamma^{k}r_{t+k+1} | s_{t} = s, a_{t} = a\right]
\end{equation}

Advantage, $\hat{A}$, is defined as the Q-value of an action in state $s$ minus the value of $s$. It can be thought of as how much better or worse an action is than the action which would be taken by the policy $\pi$.

\begin{equation}
\label{advantage}
\hat{A} = Q_{\pi}(s,a) - V_{\pi}(s)
\end{equation}

In PPO, a dual neural network setup is used. One is a policy network known as the actor and the other acts as an approximator of the value function, known as the critic. These networks share parameters, $\theta$. The policy network outputs the probability of taking actions, in a given state.

Let $z$ be the ratio between the output of an updated policy network and an old policy network, (\ref{z}).

\begin{equation}
\label{z}
z_{t}(\theta) = \frac{\pi_{\theta} (a_{t} | s_{t})} {\pi_{\theta old} (a_{t} | s_{t})}
\end{equation}

The actor objective function of PPO can then be stated, (\ref{ob_actor}), where $\epsilon$ is a hyperparameter known as the clipping range.

\begin{multline}
\label{ob_actor}
L_{t}^{ACTOR}(\theta) = \mathbb{E}_{\pi} \left[ min(z_{t}(\theta) \hat{A}_{t},\right.\\ 
\left. clip(z_{t}(\theta), 1 - \epsilon, 1 + \epsilon )\hat{A}_{t} \right]
\end{multline}

Maximising this objective function results in better actions becoming more likely and vice versa for worse actions. The minimum and clipping operators in the function prevent excessively large changes in the policy based on a single batch and performs corrections.

Finally, the overall objective function that the algorithm maximises includes this objective function and two additional terms, (\ref{loss_ppo}).

\begin{equation}
\label{loss_ppo}
L_{t}^{PPO}(\theta) = \mathbb{E}_{t} \left[ L_{t}^{ACTOR}(\theta) - c_{1}L_{t}^{VF} + c_{2}E[\pi_{\theta}](s_{t}) \right]
\end{equation}

The first additional term represents the error of the value function approximator and the second additional term encourages exploration of the environment.

Algorithm \ref{ppo_alg} demonstrates the PPO learning process.

\begin{algorithm}[H]
\caption{PPO Clip, Actor-Critic Style}\label{ppo_alg}
\begin{algorithmic}
\STATE 
\STATE Initialise values of $\theta$
\STATE \textbf{for} iteration = 1, 2, . . . \textbf{do}
\STATE \hspace{0.5cm} Run policy $\pi_{\theta old}$ in environment and collect experience
\STATE \hspace{0.5cm} Compute advantage estimates using the critic and the 
\STATE \hspace{0.5cm} Q-values found
\STATE \hspace{0.5cm} Optimize $L_{t}^{PPO}(\theta)$ with respect to $\theta$, for a given 
\STATE \hspace{0.5cm} number of epochs and a given batch size
\STATE \hspace{0.5cm}$ \theta _{old} \gets  \theta$
\STATE \textbf{end for}
\end{algorithmic}
\label{alg1}
\end{algorithm}

\section{Methodology}
This section is organised as follows. First, the simulation road network shall be explained, then the state space and action space shall be defined and finally, the training architecture shall be described.

Note that modelling the problem as an MDP is an approximation as real-world driving is not an MDP. Factors which are required to make an optimal decision such as driver intention are not known to the ego vehicle within the simulated environment.

\subsection{Road network}
The road network consists of a two-lane highway with a taper-style and parallel-style on-ramp. The taper-style ramp measures 75 m and is attached to a parallel-style ramp to the right of the main highway. The parallel-style ramp (merging lane) is 200 m long and is enclosed by two 150 m sections of highway. An example of this parallel-style and taper-style road is shown in \figurename \ref{ramp_types}.

During training and evaluation human vehicles  following the intelligent driver model described in \cite{Treiber2000} flow into the network. During training the probability of a vehicle entering the left or right lane each second is 0.1 and 0.3 respectively. This results in approximate inflows of 360 vehicles/hr in the left lane and 1080 vehicles/hr in the right lane. During evaluation these probabilities are varied.

All human controlled vehicles enter the network at 26 m/s and have desired speeds with a deviation of 0.1 m/s and a mean of 26 m/s. Human vehicles will accelerate to attain their desired speed provided they are not being blocked by another vehicle and are not yielding to give way to the merging vehicle. These vehicles are free to change lanes.

Of the human vehicles in the right highway lane, 50\% will behave in an uncooperative manner during training. This falls to 25\% during evaluation. This, along with the variation in desired speeds ensures a range of driving styles are produced.

When no ego vehicle exists in the network, one enters the on-ramp at a speed of 13 m/s. The ego vehicle is then controlled until merging is complete at which point control of the vehicle is given to the simulator and a new vehicle enters.

\subsection{Action space}
The action space, $A$, consists of a discrete set of accelerations in the range of ±3 m/s\textsuperscript{2} at 0.5 m/s\textsuperscript{2} intervals along with a discrete lane change action. The lane change action has no effect before the merging section but is still available to be selected by the agent and can only be chosen once within the merging area. The ego vehicle may begin a lane change at any point on the parallel-style ramp up to 5 m before the merging junction begins. Equation (\ref{action_list}) represents the action space.

\begin{multline}
A = \left[-3.0, -2.5, -2.0, -1.5, -1.0, -0.5, 0.0,\right.\\ \left.0.5, 1.0, 1.5, 2.0, 2.5, 3.0, \text{change lane} \right]
\label{action_list}
\end{multline}

Acceleration is limited to values in a reasonable range. As a discrete action space is used, the ego vehicle may choose only one item from (\ref{action_list}) in each timestep. Values between the listed items are not possible. The use of a discrete action space allows for an overall simplification of the action space which would otherwise require more than 1 set.

\subsection{State space}
A bounded box state space with a single row is used. The state space consists of gaps between vehicles, vehicle velocities, the ego vehicle’s longitudinal position relative to the merging point, the ego vehicle’s lateral position relative to the centre of the lane, the number of lanes in the ego vehicle’s current section of the road network and the index of the ego vehicle's current lane with 0 being the rightmost lane.

The gaps between the ego vehicle and the vehicles immediately leading and trailing it in the adjacent highway lane are included, along with their velocities. In addition to this, the gap between the leading vehicle in the adjacent lane and its leading vehicle is also recorded, along with this vehicle’s velocity. The same procedure is followed for trailing vehicles. If there is a vehicle perfectly adjacent to the ego vehicle its velocity is included.

It is assumed that the ego vehicle can observe the appropriate state values for any vehicle within the network.

\begin{figure}[!t]
\centering
\includegraphics[width=\linewidth]{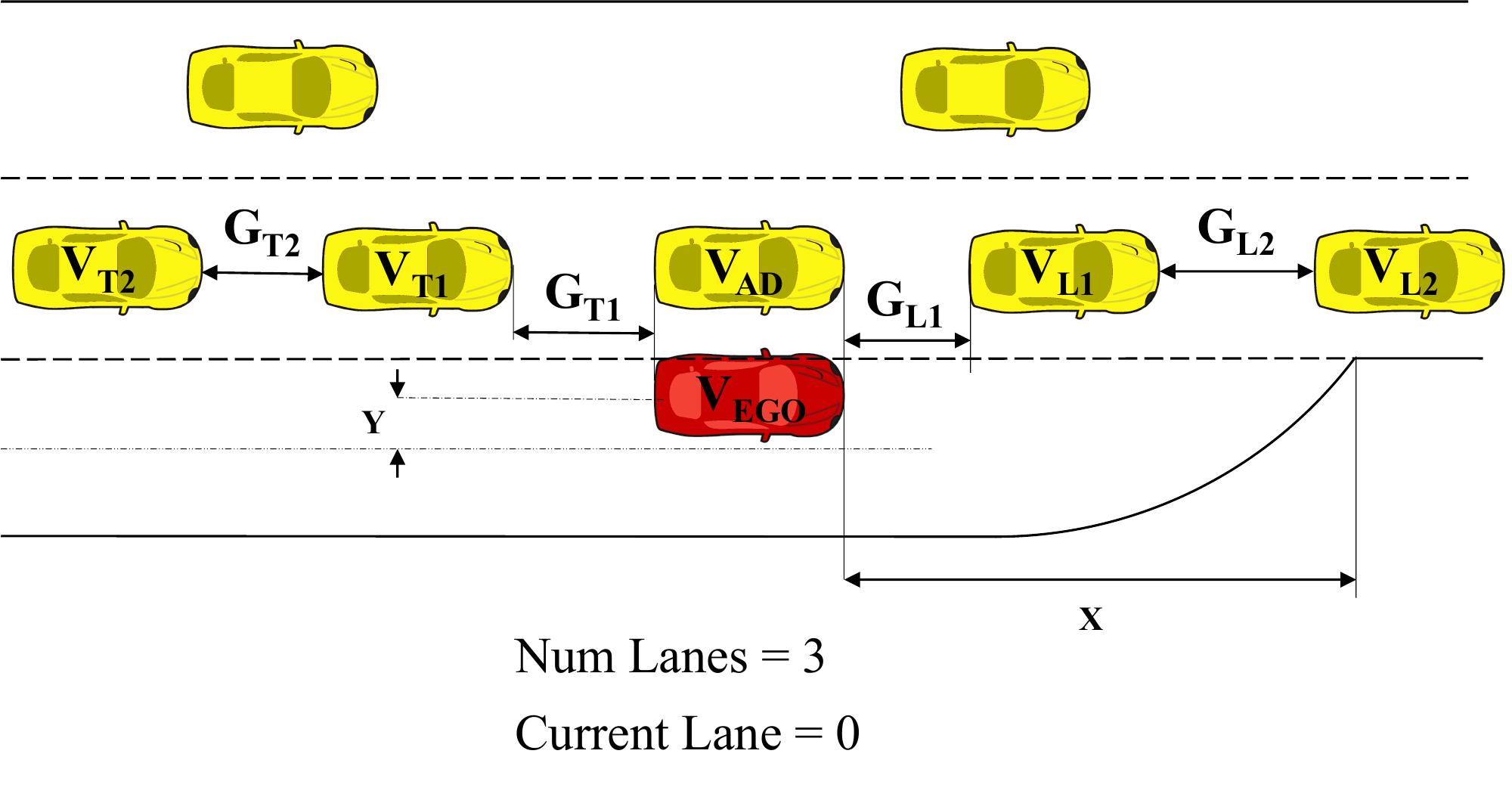}
\caption{A visualisation of the environment observation space used. Human controlled vehicles are shown in red and the ego vehicle is shown in yellow. V represents a velocity and G represents a gap. X is the longitudinal distance to the end of the merging lane and Y is the vehicle's distance from the centre of the lane.}
\label{state_space}
\end{figure}

The state at time $t$ is given by (\ref{state_list}) and visualized in \figurename ~\ref{state_space} where $V$ is a velocity, $G$  is a gap, $X$  and $Y$ are ego vehicle positional values, $C$  is the index of the ego vehicle’s current lane and $N$  is the number of available lanes.

\begin{multline}
s_{t} = \left[V_{EGO},V_{T1},V_{T2},V_{L1},V_{L2},V_{AD},\right.\\ \left.G_{T1},G_{T2}, G_{L1},G_{L2},X,Y,C,N \right]
\label{state_list}
\end{multline}

If fewer than two leading and trailing vehicles can be found, the relevant observations are set to 0. The same procedure applies to the adjacent vehicle velocity observation when there is no adjacent vehicle. When merging has occurred, the final observation takes the same values but within the ego vehicle’s current lane, not the lane adjacent to it.

All distances and speeds are normalised using appropriate values to ensure they are not exceptionally large and do not show large magnitudes of variation.

\subsection{Reward function}
\begin{figure}[!t]
\centering
\includegraphics[width=2.5in]{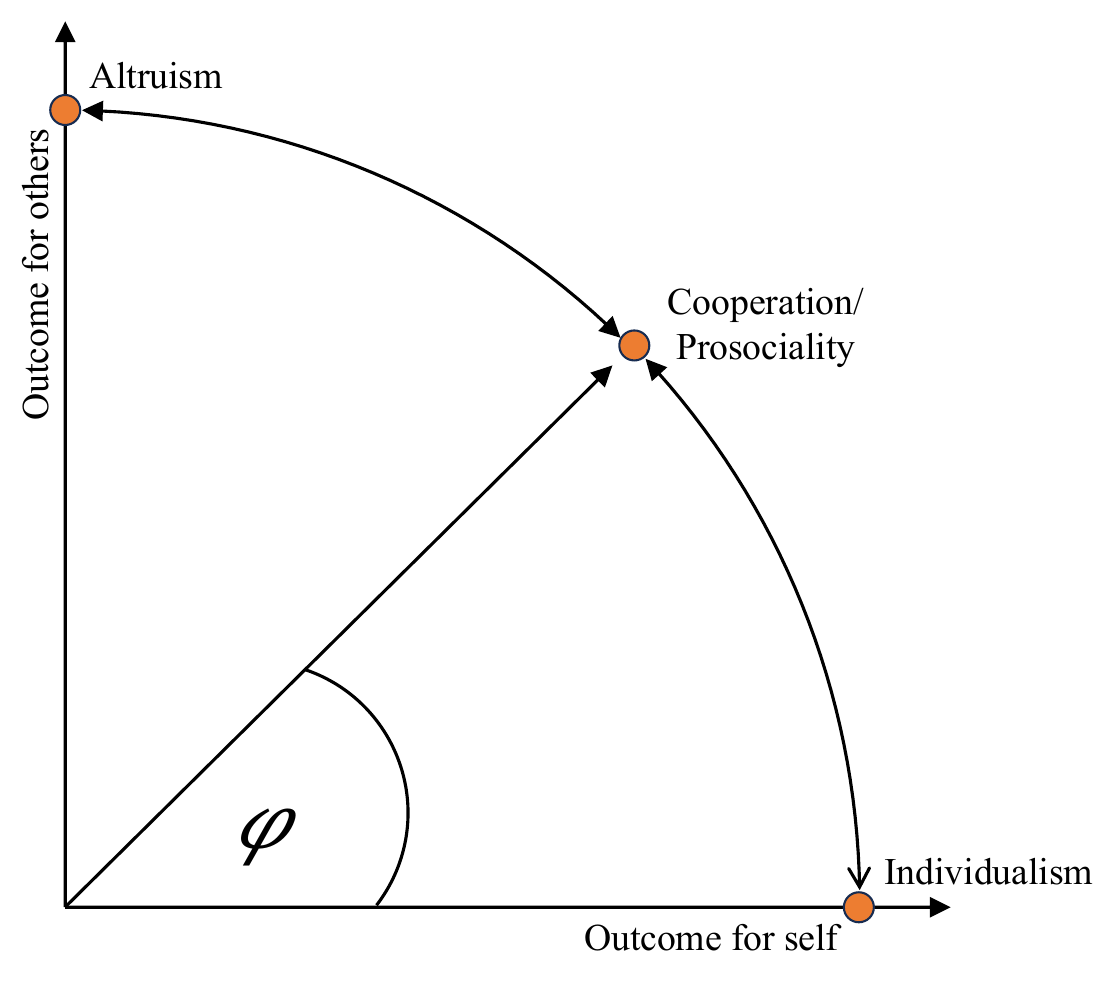}
\caption{The social value orientation ring quadrant to be used within the reward function.}
\label{svo_definition}
\end{figure}

The reward function is defined with the concept of SVO from social psychology in mind in a manner similar to Crosato et al. \cite{crosato2021, crosato2023}. SVO is defined geometrically by Griesinger et al. as the angle of a straight line from the positive x-axis in a cartesian coordinate system \cite{Griesinger1973}. A representation of this, limited to [0, $\frac{\pi}{2}$]  is shown in \figurename ~\ref{svo_definition} where the angle $\varphi$ is SVO.

In this system, as the SVO decreases from $\frac{\pi}{2}$ rad towards 0 rad, self is favoured increasingly, and others are favoured decreasingly.

The reward function is dependent upon the utility, $U$, to both the ego vehicle and the vehicles which immediately lead and follow it in the adjacent highway lane (SVs). Utilities are calculated according to (\ref{u_ego}) and (\ref{u_sv}) respectively. Weights, $w$, are applied.

\begin{gather}
U_{EGO} = w_{1} * V_{EGO} + w_{2} * min(V_{L1} - V_{EGO}, 0) \label{u_ego}\\
U_{SV} = w_{3} * G_{0} - w_{4} * G_{c} + w_{5} * min(V_{EGO} - V_{T1}, 0) \label{u_sv}
\end{gather}

Where $G_{0}$  is the chosen merging gap size and $G_{C}$ is the longitudinal distance the ego vehicle is away from the centre of the gap.

Equation (\ref{u_ego}) defines the utility to the ego vehicle in merging and assumes that the ego vehicle wishes to have a high velocity but a velocity no higher than its leading vehicle as this could require it to brake when merging is complete, disrupting the smoothness of its travel and potentially leading to a collision.

Equation (\ref{u_sv}) defines the utility to the SVs in merging and assumes that the SVs prefer the ego vehicle to merge into larger gaps over smaller ones, into the centre of a gap as opposed to at the edge and at a speed no less than that of the immediately trailing vehicle. This is to ensure appropriate spacing between vehicles and to avoid SV braking.

If no leading or trailing vehicle can be found, it is assumed there is one just outside the network, travelling at the same speed as the ego vehicle.

For head and tail gaps greater than some distance, $d$, $G_{c}$ is classed as 0. This is because when the head and tail gaps are sufficiently large, the ego vehicle’s distance from the centre has a negligible effect on the utility to SVs provided the other objectives of SV utility are met. If the ego vehicle has merged with large enough head and tail gaps, its proximity to the centre of the merging gap is irrelevant, provided it is travelling at an appropriate velocity.

The reward function is given by (\ref{reward_fn}) where $C$ is a constant. This function contains both the ego vehicle and SV utility. The values used in this report are listed in Table \ref{reward_params}.

\begin{equation}
\label{reward_fn}
  r =
    \begin{cases}
      0 & \text{before merging zone}\\
      C & \text{if crashed}\\
      U_{EGO} cos(\varphi) + U_{SV} sin(\varphi) & \text{otherwise}
    \end{cases}       
\end{equation}

Adjusting the SVO varies the degree to which the agent prioritises the utility to other road users.

\begin{table}[!t]
\renewcommand{\arraystretch}{2}
\caption{Reward function parameters\label{reward_params}}
\centering
\begin{tabular}{c|c|c|c|c|c|c|c|c}
\hline
Param & $\varphi$ & $w_{1}$ & $w_{2}$ & $w_{3}$ & $w_{4}$ & $w_{5}$ & $d$ & $C$\\
\hline
Value & $\frac{\pi}{4}$ & $\frac{1}{13}$ & $\frac{4}{13}$ & $\frac{15}{389}$ & $\frac{6}{13}$ & $\frac{8}{13}$ & $40$ & $-20$\\
\hline
\end{tabular}
\end{table}

\subsection{Training architecture}
\begin{figure}[!t]
\centering
\includegraphics[width=2.5in]{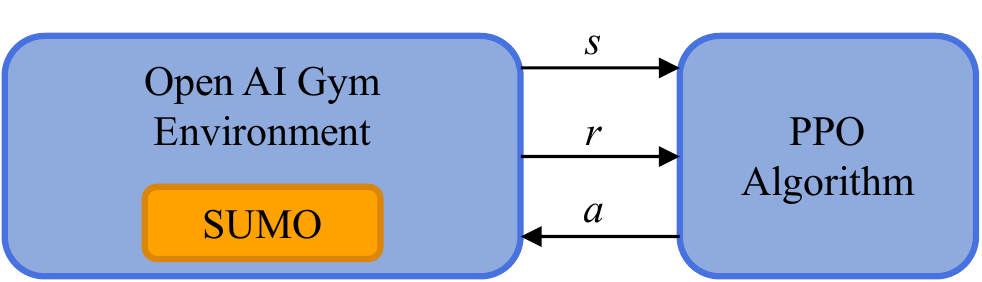}
\caption{The architecture of the software used for training the reinforcement learning model.} \label{architecture}
\end{figure}

The reinforcement learning environment was constructed using OpenAI Gym \cite{brockman2016} coupled with the SUMO simulator in Python. SUMO is a microscopic traffic simulator that provides a suite of tools for simulating and analysing traffic situations \cite{SUMO2018}. Vehicles in SUMO can be controlled by the simulator according to some driving model or they can be controlled remotely. The TraCI API was used to interact with the simulator, attaining states from the simulator and sending actions to the ego vehicle. States were sampled and actions were chosen at a rate of 10 Hz, thus each timestep was a simulated 0.1 s. The architecture used is shown in \figurename ~\ref{architecture}.

A new random seed was used every time the simulator was reset due to a collision or timeout. The random seed controls the timing of vehicle spawning and thus this prevented the model from being overfitted to any one vehicle flow pattern. Timeout occurred after a simulated 150 s.

The agent was trained using an on-policy algorithm, PPO and an off-policy algorithm, DQN. As the PPO algorithm showed greater success, this will be focused upon. A wider range of algorithms has not been evaluated as our focus is to demonstrate the novel reward function and the necessity of SV consideration in reward function design. The agent was trained for 15 million timesteps using the PPO implementation in Stable Baselines3 \cite{raffin2021} on an Intel Xeon E3-1240 V6 and an NVIDIA Quadro P2000. It should be noted that this implementation varies slightly from the canonical implementation in \cite{Schulman2017}. Training data was collected in 20 parallel environments and GPU acceleration was used with the goal of decreasing training time. During training the model was evaluated periodically over 50 episodes in an environment with an overall inflow rate identical to that of the training network but with the 75\% - 25\% cooperative to uncooperative split.

The neural networks used had two hidden layers of width 64. The width of the input and output layers correspond to the number of items in the state and action sets respectively. The rectified linear unit activation function described in \cite{goodfellow2016} was used. The default algorithm hyperparameters provided by Stable Baselines3 were used. Key hyperparameter values for the PPO algorithm are given in Table \ref{ppo_hyperparam}.

\begin{table}[!t]
\caption{Learning agent hyperparameters\label{ppo_hyperparam}}
\renewcommand{\arraystretch}{1.5}
\centering
\begin{tabular}{c|c}
\hline
Hyperparameter & Value\\
\hline
Learning rate & 0.0003\\
\hline
Horizon & 2048\\
\hline
Minibatch size & 64\\
\hline
Number of epochs & 10\\
\hline
Discount factor & 0.99\\
\hline
Clipping range & 0.2\\
\hline
Value function coefficient & 0.5\\
\hline
Entropy coefficient & 0\\
\hline
\end{tabular}
\end{table}

\section{Evaluation}

\subsection{Training}
\begin{figure}[!t]
\begin{minipage}[c]{0.49\linewidth}
\includegraphics[width=\linewidth]{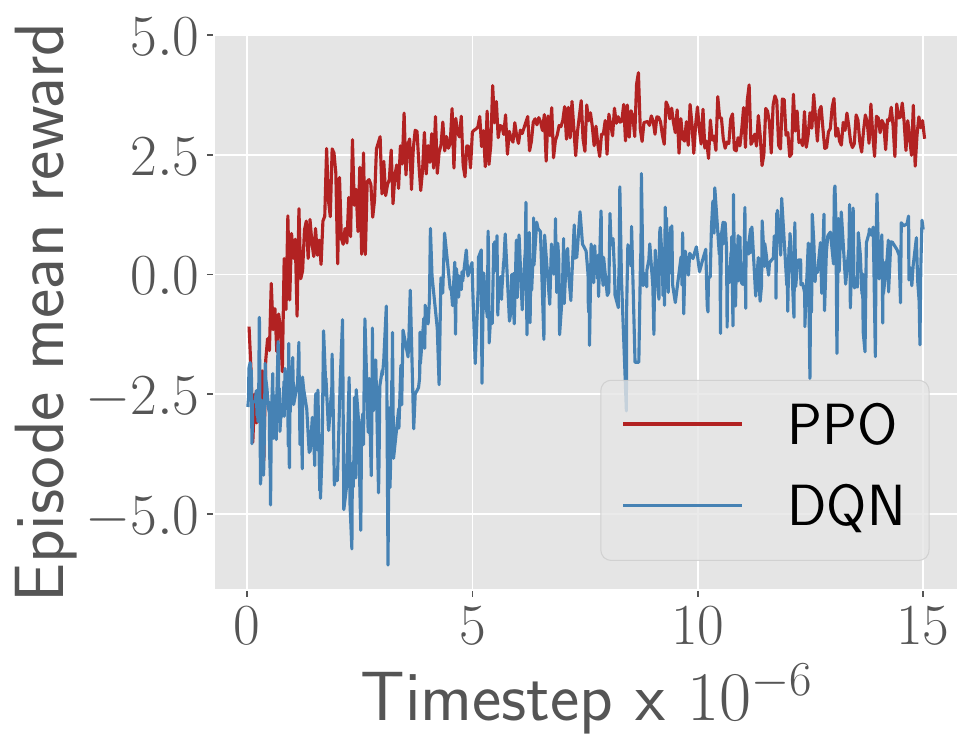}
\caption{The episode mean reward during training. The progress of the PPO algorithm is shown in red while the DQN algorithm is shown in blue.} \label{reward}
\end{minipage}
\hfill
\begin{minipage}[c]{0.49\linewidth}
\includegraphics[width=\linewidth]{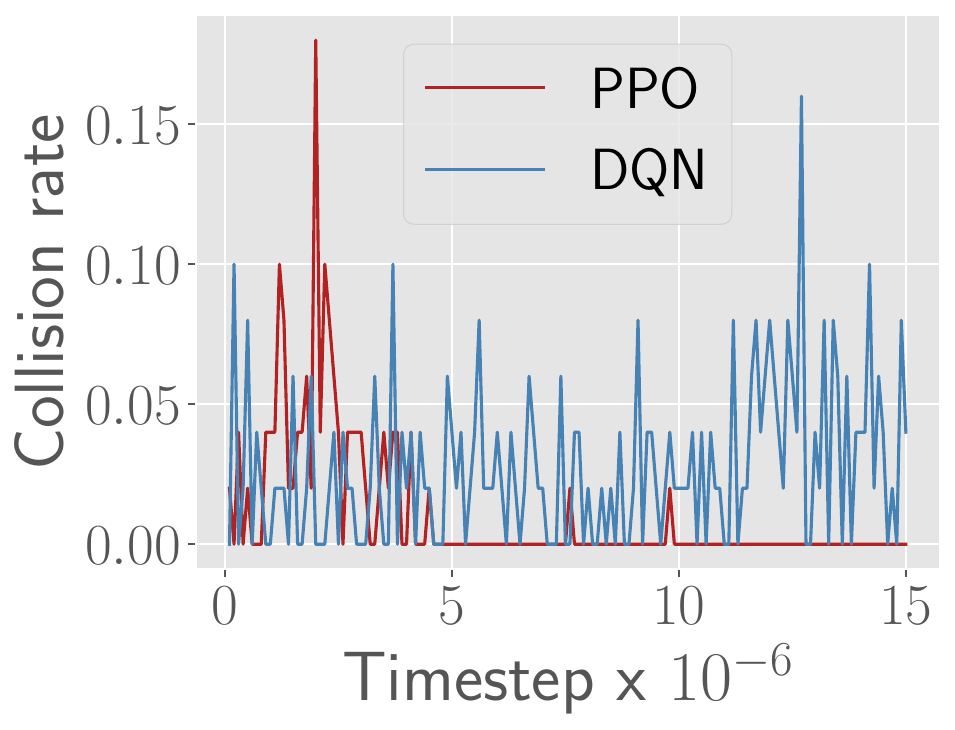}
\caption{The percentage of collisions within the evaluation set during training. The progress of the PPO algorithm is shown in red while the DQN algorithm is shown in blue.} \label{collisions}
\end{minipage}%
\end{figure}

The model was found to converge successfully for all values of SVO using both the PPO and DQN algorithms. The average episode reward attained during each roll out in training for an SVO value of $\frac{\pi}{4}$ is shown in \figurename ~\ref{reward}. It can be seen that the reward value starts around 0 before arriving at a quasi-steady value of approximately 3 around 10 million timesteps for the PPO implementation. The collision rate shows a high level of variability in the early stages of training before falling to 0 at 7.7 million timesteps and remaining there until the completion of training, \figurename ~\ref{collisions}. The model’s training finished after 15 million timesteps. It can be seen that the DQN algorithm fails to achieve a steady 0\% collision rate. Due to the success of PPO, this algorithm is used for further analysis of the proposed reward function. A similar result would be expected over the range of SVO values.

\subsection{Experiments}
The produced models were evaluated in the simulated environment. Each evaluation took place over 100 merges. Traffic densities used are given in Table \ref{mixed_density_results}. Evaluations in medium traffic mode were undertaken for a range of SVO values to indicate the importance of each of the two reward function elements and the effect of SVO variance. Results are shown in Table \ref{svo_results}. The most successful SVO value underwent further evaluation in a number of other traffic densities, Table \ref{prosocial_results}.

\begin{figure*}[!t]
    \includegraphics[width=\textwidth]{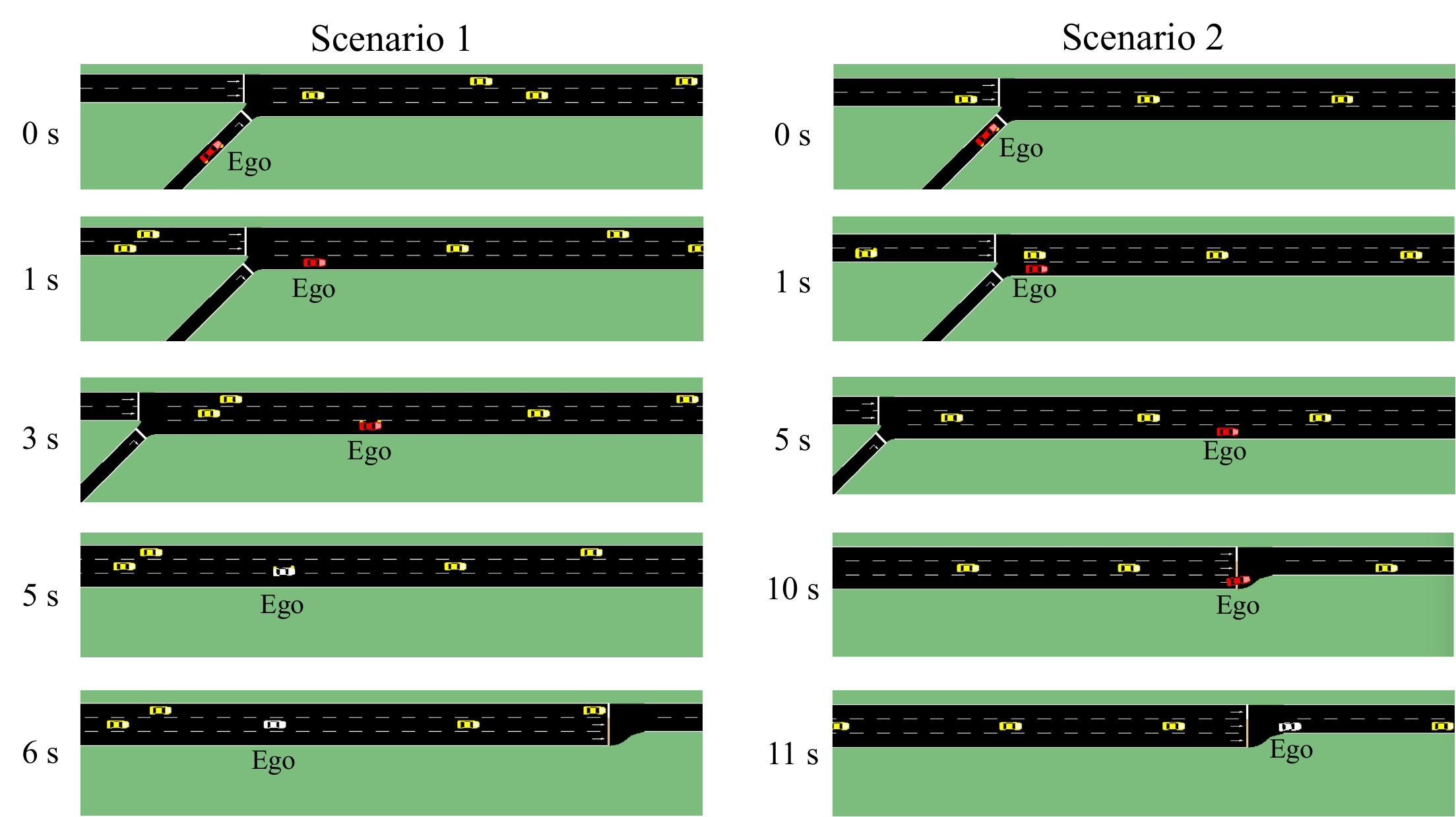}
    \caption{Two example merges. Human controlled vehicles are shown in yellow. The ego vehicle is shown in red. Once the centre of the ego vehicle has entered the adjacent lane, the ego vehicle changes to white and control is given to the simulator. In the first example, the ego vehicle merges shortly after the taper-style lane ends. In the second, the ego vehicle yields to an oncoming vehicle and performs gap following, merging at the end of the parallel-style section.}
    \label{example_merges}
\end{figure*}

A number of example merges with the central SVO value are shown in \figurename ~\ref{example_merges}. In the first of these, the ego vehicle finds a gap immediately after completing the taper-style ramp. The ego vehicle achieves appropriate gap positioning and speed before initiating a merge. In the second, the ego vehicle arrives at the parallel-style ramp and yields to the vehicle in the adjacent lane. The ego vehicle positions itself appropriately in the gap behind this vehicle at an appropriate speed before merging at the end of the lane.

The TTC with the first trailing and first leading vehicles are calculated according to (\ref{TTCT1}) and (\ref{TTCL1}) respectively when merging occurs. These metrics take into account head and tail gap size however these are not considered in isolation as what defines acceptable gaps is dependent upon the wider context of the merge.

\begin{gather}
TTC_{T1} = \frac{G_{T1}}{V_{T1} - V_{EGO}} \label{TTCT1}\\
TTC_{L1} = \frac{G_{L1}}{V_{EGO} - V_{L1}} \label{TTCL1}
\end{gather}

According to these definitions, if the output values are equal to or less than 0, the ego vehicle has ensured that a crash would never occur under constant velocity. The percentage of merges in the test set with a TTC of less than 10 s is listed in Tables \ref{svo_results}, \ref{prosocial_results}. A lower number indicates more aggressive merging behaviour towards the leading or trailing vehicle.

The vehicle velocity was recorded at the point of merging. This average velocity is also listed in Tables \ref{svo_results}, \ref{prosocial_results}. A conflict is when hard braking is enacted by at least one party.

The ratio of $G_c$ to $G_0$ indicates the fraction of the overall gap the vehicle is displaced from the gap centre by. A ratio is used as the centrality of the vehicle is more critical in smaller gaps and thus absolute values are unclear. The percentage of merges where the ratio is greater than 0.5 is listed in Tables \ref{svo_results}, \ref{prosocial_results}. A higher number indicates poorer gap positioning.

\begin{figure}[!t]
    \centering
    \includegraphics[width=0.49\linewidth]{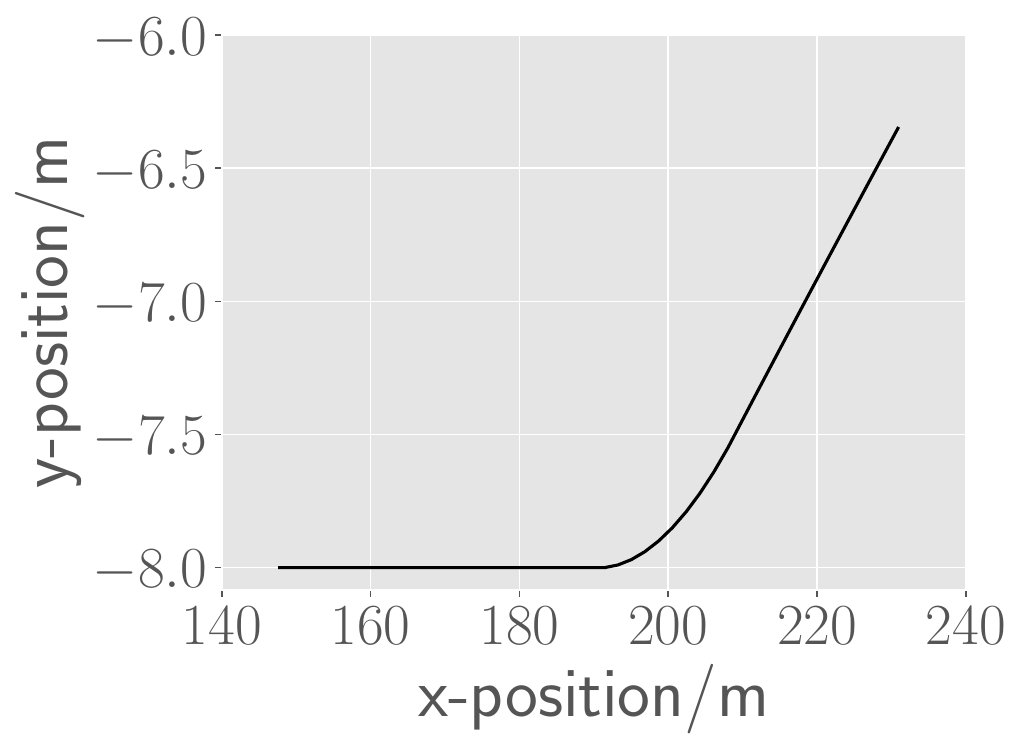}
    \includegraphics[width=0.49\linewidth]{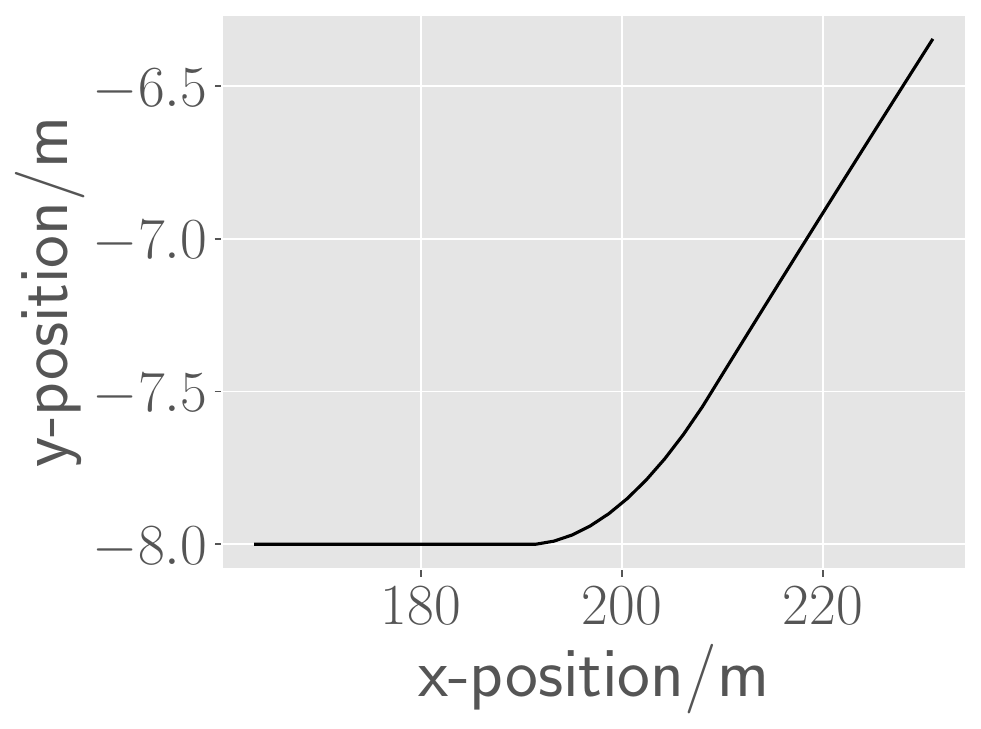}
    \caption{An example vehicle merging trajectory from each experiment is plotted. A smooth S motion is produced by the SUMO simulator.}
    \label{example_trajectories}
\end{figure}

Example merging trajectories are plotted in \figurename ~\ref{example_trajectories}.

\begin{table}[!t]
\caption{Traffic densities used for model testing\label{mixed_density_results}}
\renewcommand{\arraystretch}{1.5}
\centering
\begin{tabular}{c|c|c|c}
\hline
 & Easy Mode & Medium Mode & Hard Mode\\
\hline
Right lane inflow / vehicle/hr & 405 & 810 & 1013 \\
\hline
Left lane inflow / vehicle/hr & 90 & 180 & 225 \\
\hline
\end{tabular}
\end{table}

\subsection{Results}
\begin{table}[!t]
\caption{A comparison of driving behaviours for varying SVO values\label{svo_results}}
\renewcommand{\arraystretch}{1.75}
\centering
\begin{tabular}{c|c|c|c}
\hline
$\varphi$ & 0 & $\mathbf{\frac{\boldsymbol{\pi}}{4}}$ & $\frac{\pi}{2}$\\
\hline
Collisions / \% & 1 & \textbf{0} & 0\\
\hline
Conflicts / \% & 9 & \textbf{2} & 7\\
\hline
Average merge velocity / m/s & 21.4 & \textbf{24.1} & 24.7\\
\hline
$TTC_{L1}$ \textless 10 s / \% & 0  & \textbf{3} & 9\\
\hline
$TTC_{T1}$ \textless 10 s / \% & 5 & \textbf{0} & 0\\
\hline
$G_c/G_0$ \textgreater 0.5 / \% & 19 & \textbf{6} & 36\\
\hline
\end{tabular}
\end{table}

\begin{table}[!t]
\caption{An evaluation of the prosocial SVO model across varied traffic conditions\label{prosocial_results}}
\renewcommand{\arraystretch}{1.5}
\centering
\begin{tabular}{c|c|c|c}
\hline
& Easy Mode & Medium Mode & Hard Mode\\
\hline
Collisions / \% & 0 & 0 & 0\\
\hline
Conflicts / \% & 2 & 2 & 5\\
\hline
Average merge velocity / m/s & 23.4 & 24.1 & 23.3\\
\hline
$TTC_{L1}$ \textless 10 s / \% & 0  & 3 & 0\\
\hline
$TTC_{T1}$ \textless 10 s / \% & 1 & 0 & 2\\
\hline
$G_c/G_0$ \textgreater 0.5 / \% & 6 & 6 & 2\\
\hline
\end{tabular}
\end{table}

At an SVO of 0 wherein SV utility was disregarded, dangerous, individualistic behaviour was shown demonstrating the requirement for SV consideration. While there was only 1 collision, the model showed aggressive behaviour with a large number of near misses, primarily due to merging in front of faster-moving vehicles in an improper gap position. 19\% of merges with this model were completed with a $G_c/D_d$ ratio of over 0.5 and the TTC with tailing vehicles was less than 10 for 5\%.

At an SVO of $\frac{\pi}{2}$, the model sought to maximise the utility to SVs in merging. This resulted in a poor ride for the ego vehicle which often merged inappropriately relative to its leading vehicle and had to brake. 9\% of merges had a TTC with the leading vehicle of less than 10 and 36\% of merges had a $G_c/G_0$ ratio of greater than 0.5. The model, while collision-free, produced 7 conflicts. 4 of these conflicts involved the ego vehicle braking hard.

The central SVO value can be seen to be the best of those surveyed at the medium traffic density, harnessing the advantages of both. It presents the fewest conflicts and no collisions while also attaining acceptable values in all other categories. The TTC with the trailing vehicle was less than 10 s in only 3\% of merges and the model's preference for gap centrality was high with 94\% of merges having a $G_c$ to $G_0$ ratio of less than 0.5. This indicates that the ego vehicle was merging in a socially acceptable manner, at an appropriate speed and gap position relative to the trailing vehicle. This constitutes positive social behaviour. Merging speeds were close to the network entry speed of 26 m/s and the TTC with the leading vehicle was never less than 10 s indicating that the ego vehicle did not have to brake upon merging. This indicates a fulfilment of the ego vehicle's goals. 

In general, the safety and social cooperation of the prosocial SVO value sustained over a range of traffic densities, Table \ref{prosocial_results}. No collisions occurred at any density and conflicts remained low. Correct merging velocities and gap positioning were attained, as shown by the low TTC and $G_c/G_0$ results which show little variance across traffic densities. Even in more dense traffic, the ego vehicle managed to attain a speed within 3 m/s of the network entry speed.

It can be seen from the results in Table \ref{svo_results} that it is important to consider the utility to both the ego vehicle and the SVs in the reward function. Of particular note is the effect of considering only the ego vehicle's goals which resulted in the most conflicts and anti-social driving behaviour. In addition, the safety and social courtesy of the central SVO model are highlighted as the results indicate it minimised instances of cutting in front or behind at an inappropriate speed or position and produced minimal conflicts as well as no collisions in a range of traffic densities.

\subsection{Comparison}
It is difficult to fully compare RACE-SM to the literature as the effect on SVs and the conflict rate are rarely reported or considered. The model can however be compared in terms of collisions, Table \ref{results_comparison}. 

\begin{table}[!t]
\caption{A comparison of the collision rate between RACE-SM and a number of baselines in the literature\label{results_comparison}}
\renewcommand{\arraystretch}{1.75}
\centering
\begin{tabular}{c|c|c}
\hline
Author & Test data set & Collision rate / \% \\
\hline
\textbf{Poots} & \textbf{Custom} & \textbf{0}\\
\hline
Triest et al. \cite{Triest2020} & NGSIM & 4.2 \\
\hline
Liu et al. \cite{Liu2023} & US101 & 0.51 \\
\hline
Lubars et al. \cite{Lubars2021} & Custom & 0\\
\hline
Lin et al. \cite{Lin2020} & Custom & 0\\
\hline
\end{tabular}
\end{table}

It can be seen that RACE-SM, either matches or surpasses those in the literature in terms of collision rate while introducing a focus on the SVs to produce behaviour that is not just collision free but also socially acceptable, avoiding conflicts and driver frustration. The importance of this addition is demonstrated in Table \ref{svo_results}.

\subsection{Limitations and Future Work}
Despite these results, a number of limitations exist. The model is optimised for moving traffic on the given road network. Future work should consider a more varied set of traffic conditions and road networks, and parameter tweaking for increased generalisation.

The accuracy of the model and the road network is limited by the accuracy of the dynamics implemented within the SUMO simulator, the default vehicle parameters provided by SUMO and the simulator’s network modelling constraints. The simulator is under continuous improvement however the model is also subject to any current bugs. Higher fidelity simulators such as CARLA are available \cite{carla} but these have significant technical overheads.

It is assumed that all vehicles within the network can be detected. Future work should more accurately consider sensing hardware restraints.

\section{Conclusions}
This work proposed a reinforcement learning based approach for social on-ramp merging in human controlled traffic which explicitly considers the utility to both the ego vehicle and its SVs. The key innovation of this work is the explicit and comprehensive consideration of the utility to SVs in merging applied within a road network populated by both cooperative and uncooperative drivers. The produced model showed a promising ability to merge in a socially courteous manner. The results indicate:
\begin{enumerate}
\item{Socially courteous behaviour can be produced through direct consideration of the utility to SVs in the reward function. This consideration resulted in the TTC with trailing vehicles upon merging being larger than 10 s for almost all merges and effective ego vehicle gap placement.}
\item{Failure to consider the goals of SVs during merging results in behaviour which may exhibit few collisions but is not socially acceptable or safe for real-world deployment.}
\item{Direct consideration of the objectives of both the ego vehicle and the SVs in the DRL model’s reward function resulted in behaviour that maximises utility to the ego vehicle within the constraints of other objectives. In medium traffic, an average merging velocity of 24.1 m/s in a network with an entry speed of 26 m/s was attained and all ego vehicles had no prompt need to decelerate.}
\item{SVO can be used within a reinforcement learning framework to effectively vary the driving style of an ego vehicle between altruistic and individualistic.}
\end{enumerate}

\bibliographystyle{IEEEtran}
\bibliography{IEEEabrv, ref}

\begin{IEEEbiography}[{\includegraphics[width=1in,height=1.25in,clip,keepaspectratio]{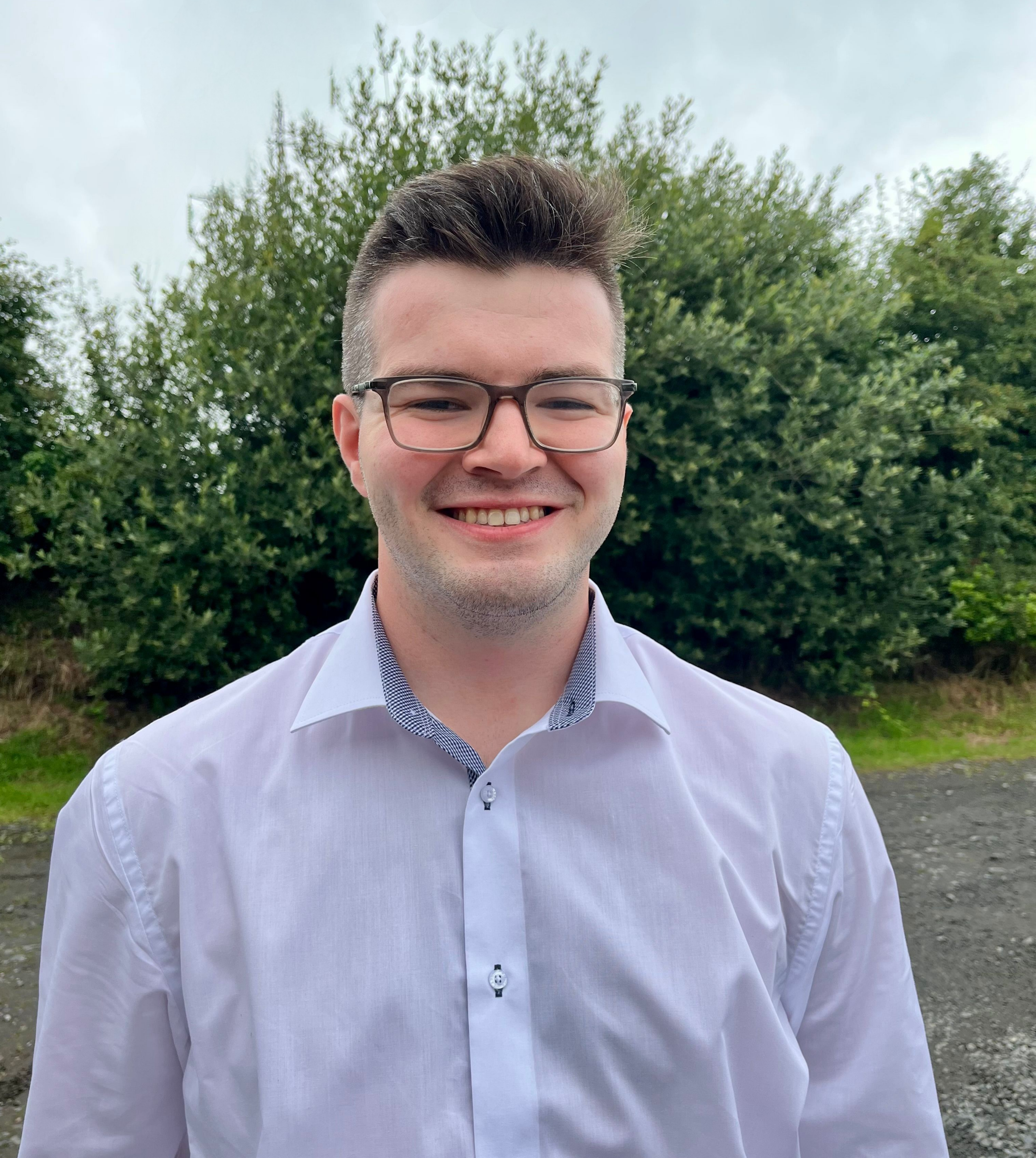}}]{Jordan Poots}
received his Bachelor's degree in Mechanical Engineering from Queen's University, Belfast in 2023. He is currently a Master's student at Queen's, studying Software Development. His working interests include AI and software engineering.
\end{IEEEbiography}

\end{document}